\documentclass[fleqn,10pt]{wlscirep}
\usepackage[utf8]{inputenc}
\usepackage[T1]{fontenc}
\usepackage{cite}

\usepackage{graphicx}%
\usepackage{rotating}
\usepackage{subcaption}
\usepackage{multirow}%
\usepackage{url}
\usepackage{amsmath,amssymb,amsfonts}%
\usepackage{amsthm}%
\usepackage{mathrsfs}%
\usepackage[title]{appendix}%
\usepackage{xcolor}%
\usepackage{textcomp}%
\usepackage{manyfoot}%
\usepackage{caption}
\usepackage{booktabs}%
\usepackage{algorithm}%
\usepackage{algorithmicx}%
\usepackage{algpseudocode}%
 \usepackage{multirow}
 \usepackage{overpic}
\usepackage{booktabs}
\usepackage{pifont} 

\usepackage{listings}%
\title{Investigating Generative AI models for Forensic Craniofacial Reconstruction}

\author[1,*]{Ravi Shankar Prasad}
\author[2]{Dinesh Singh}

\affil[1,2]{Visual Intelligence and Machine Learning (VIML) Group, School of Computing and Electrical Engineering, Indian Institute of Technology Mandi, Mandi, Himachal Pradesh 175005, India}

\affil[*]{d23033@students.iitmandi.ac.in}

\affil[1,2]{these authors contributed equally to this work}

\begin{abstract}
Craniofacial reconstruction in forensics is one of the processes to identify victims of crime and natural disasters. Identifying an individual from their remains plays a crucial role when all other identification methods fail. Traditional methods for this task, such as clay-based craniofacial reconstruction, require expert domain knowledge and are a time-consuming process. At the same time, other probabilistic generative models like the statistical shape model or the Basel face model fail to capture the skull and face cross-domain attributes. Looking at these limitations, we propose a generic framework for craniofacial reconstruction from 2D X-ray images. Here, we used various generative models (i.e., CycleGANs, cGANs, etc) and fine-tune the generator and discriminator parts to generate more realistic images in two distinct domains, which are the skull and face of an individual. This is the first time where 2D X-rays are being used as a representation of the skull by generative models for craniofacial reconstruction. We have evaluated the quality of generated faces using FID, IS, and SSIM scores. Finally, we have proposed a retrieval framework where the query is the generated face image and the gallery is the database of real faces. By experimental results, we have found that these generative models can be used as an assisting tool for craniofacial identifications in forensic science.
\end{abstract}
\begin{document}

\flushbottom
\maketitle
%
%
\thispagestyle{empty}


\section{Introduction}
Craniofacial reconstruction aims to identify individuals based on the structure of their underlying skulls. This technique is not only utilised in the field of forensics but also in various fields, including history, anthropology, and archaeology. This method is also helpful in forensic cases, such as identifying terrorists and victims of large landslides in mountainous regions or tsunamis in coastal areas. 
When an unidentified skull is found at a crime scene or during a natural disaster, the forensic team tries to find the identity of the deceased individual. But, the identification is difficult when other means of identification are not possible (e.g., soft tissue, hair strands) because over time, soft tissue and hair strand decomposes. To identify the individual based on the DNA is also very difficult because the forensic team does not know with whom they have to match the DNA samples. Then, at the last, the forensic team tries to reconstruct the face by applying clay over the skull. After reconstructing the face from the given skull, they compare this reconstructed face with a database of facial images of missing persons to identify the individual. However, this method is time-consuming and requires an artist with expert knowledge in anthropology.
\begin{figure}[!ht]
        \centering
        \includegraphics[width=1\linewidth, keepaspectratio,trim={4.8cm 4.5cm 7.8cm 3.5cm},clip]{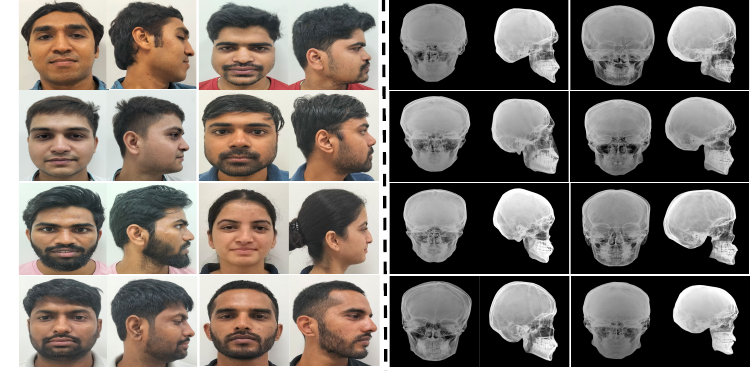}
        \caption{Sample images from the \emph{S2F} dataset, show paired images of face and skull from front and lateral side, respectively.}
        \label{fig:dataset}
    \end{figure}
In the traditional method of craniofacial reconstruction, a moldable material is superimposed by hand onto an unidentified skull based on the soft tissue information \cite{hwang2012facial} and other anatomical data. The detailed study by~\cite{wilkinson2010facial} shows that this process is highly subjective and requires a significant amount of creative interpretation, which can sometimes lead to unpredictable results. 

To overcome the problem faced by traditional method, computerized automatic methods have been developed for skull-face overlay. There are primarily two main methods used in computer-aided craniofacial identification process. The first method is craniofacial superimposition ~\cite{damas2011forensic, campomanes2014computer}, which directly superimposes a 2D image of the skull onto an image of the face. In the second method ~\cite{claes2010computerized}, reconstruction and editing of the face are done using a 3D statistical shape model~\cite{dai2020statistical}. However, processing 3D skull and face data is not a trivial task, as it involves several scans at various depths and takes higher time and computational costs. Hence, to solve this problem, we used the 2D X-ray scans of the face, which are easier to obtain compared to the 3D CT image and leading to faster results and reduced cost. Recent advances in machine learning and deep learning models, particularly in the field of generative models like GANs~\cite{goodfellow2020generative}, have made the task of image generation more popular. Various improved versions of GANs~\cite{zhu2017unpaired,park2020contrastive} have introduced significant improvements in image translation. But, this image translation is done in publicly available dataset (i.e., converting horse image to zebra image or converting cat image to dog image) and also application of these generative models are limited to forensic craniofaical reconstruction due to unavailability of publically available benchmark dataset. Hence, in this work, we propose a craniofacial reconstruction based on GenAI to synthesise craniofacial images from 2D X-ray scans of the face. This approach can be redefined as an image-to-image translation task, to generate corresponding facial images from 2D skull images. Our approach has two main advantages. First, it effectively captures the complex relationship between the skull and the face. Second, unlike CT scans, it does not require data from scans of the entire head, as it only uses frontal and lateral views from facial X-ray scans. The overall structure of the paper is organized as follows: Section 2 reviews related work in craniofacial identification and generative modeling. Section 3 contains dataset descriptions. Section 4 describes the proposed methodology, which also includes data preprocessing and model architectures. Section 5 presents the experimental setup, evaluation metrics and results. Section 6 discusses the limitation, challenges and future work of the study followed by the conclusion.

The major contributions of this research are as follows;
\begin{itemize}

    \item We have proposed a generative framework for 2D craniofacial reconstruction using different generative AI models. 
    \item We have also extended the benchmark \emph{S2F} dataset for this purpose. The details of the dataset are given in section~\ref{sec:dataset}. Figure~\ref{fig:dataset} shows a sample of the dataset.
    \item We conducted extensive experiments and comparative analysis on various generative models (i.e., CycleGANs, cGANs) for cross-domain craniofacial reconstruction and evaluated the quality of generated face images using FID, IS and SSIM scores.
    \item We also evaluated the performance of different deep models (i.e., ResNet, VGG16, DenseNet) on face retrieval tasks using two important retrieval metrics, Recall@k and mean Average Precision (mAP@k).
\end{itemize}

\section{Related work}

\textbf{Craniofacial reconstruction.} A few attempts have been made in craniofacial reconstruction (CR) and matching the cranio (i.e., skull), to the face. Traditional methods,~\cite{wilkinson2010facial, hwang2012facial} uses anatomical knowledge and soft tissue depth information to describes how these information are used for reconstructing a face from the given skull. Apart from manual intervention and expert dependency, these methods also have limited scalability, lack generalisation across populations, and possess limited ground truth validation.
In classical machine learning methods, there have been a few works that has been explored for CR, such as least squares support vector regression~\cite{li2014craniofacial}, latent root regression~\cite{berar2011craniofacial}, and partial least squares regression~\cite{jia2021craniofacial}. However, these methods struggle with limited data, often miss local facial details, and depend heavily on accurate landmarks annotations.  These limitations highlight the necessity for more adaptable, data-driven deep learning methods that effectively capture complex patterns in craniofacial data. As the primary goal of CR is to identify the individual either by directly matching skull with the face or by reconstructing face from the given skull. However, in direct skull-to-face matching and to learn cross-domain representations to identify individuals from the X-ray scans (i.e., skull),~\cite{prasad2025cross} attempt to match the embeddings of skull images against the gallery face images and identify which person's skull it is. But, due to the large gap between the two modalities, which is the  skull and face, models are overburdened with the responsibility of aligning the two modalities. Hence, our approach tackles this problem by converting the skull image into a face image and then matching the images in a similar domain and modalities. Also, the reconstruction can be used for the identification of individuals by humans.

\textbf{ Generative AI.} With the advancement of deep models and generative artificial intelligence (GenAI), the task of generating images has become very popular. Generative adversarial networks (GANs) have demonstrated strong performance in generating images in unpaired image-to-image translation tasks, particularly with the introduction of CycleGAN  ~\cite{zhu2017unpaired}. CycleGAN eliminates the need for paired data by enforcing cycle consistency loss, allowing effective domain translation between two unaligned datasets. Few works were conducted on craniofacial reconstruction task using GANs~\cite{ pengyue2021cfr,li2022cr, zhang2022end, zhao2024intrinsic}. However, these works utilises a CT scan datasets, which are more challenging to obtain than X-ray scans because of very high cost, high compute and high radiation exposure. We have also observed that these GenAI models often struggle to preserve fine-grained semantic information across domains, particularly in tasks involving structural complexity, such as skull-to-face synthesis. To address this issue, the patch-wise contrastive loss, proposed in ~\cite{park2020contrastive}, enforces consistency in patch-wise semantic relations between the input and generated output. Hence, these methods improve the overall structural alignment in mapping cross-domain translation.
In our work, we have employed multiple methods inspired by these papers; however, instead of utilising generic and CT scan datasets, we have used a benchmark \emph{S2F} dataset. Also, the previous models computed only the FID scores~\cite{jayasumana2024rethinking} as a generative evaluation metric; we have included Inception Score (IS ) ~\cite{barratt2018note} and Structural Similarity Index scores (SSIM) ~\cite{nilsson2020understanding}, Learned Perceptual Image Patch Similarity (LPIPS)~\cite{zhang2018unreasonable} and ArcFace similarity~\cite{deng2019arcface} as quantitative metrics to evaluate our generated images.

\textbf{Image translation and cycle-consistency.} Basically, image translation means mapping an image from one domain to another domain while preserving underlying structure. In paired settings, this is typically achieved using adversarial and reconstruction losses~\cite{ goodfellow2014generative,isola2017image}, In unpaired settings, cycle consistency has become a widely adopted constraint~\cite{zhu2017unpaired,yi2017dualgan,kim2017learning}, enforcing that a translated image can be mapped back to its original domain. Alternative frameworks which are done in UNIT~\cite{liu2017unsupervised} and MUNIT~\cite{huang2018multimodal} propose learning a shared latent representation between domains, which enables more flexible cross-domain translation. Recent works have further extended these methods to multi-domain and multi-modal scenarios~\cite{choi2018stargan,almahairi2018augmented,zhu2017toward,lee2018diverse,liu2019few}, improving visual quality and diversity~\cite{tang2019attention,zhang2019harmonic,gokaslan2018improving,liang2018generative,wu2019transgaga}. However, the strict bijective assumption imposed by cycle consistency can be limiting, particularly in tasks like skull-to-face translation, where information loss and ambiguity are inherent.
Despite these advances, the application of generative models to craniofacial reconstruction using 2D X-ray images remains unexplored. The task of CR posses significant challenges due to limited availability of benchmark skull-to-face datasets and the complexity of cross-domain identity preservation. Hence, our work addresses this gap by systematically evaluating multiple generative approaches for skull-to-face translation using a curated 2D X-ray S2F dataset.

\section{Dataset}\label{sec:dataset}
We have used the \emph{S2F} dataset ~\cite{prasad2025cross} and extended the dataset samples from 40 individuals to 51 individuals following the same dataset creation procedures and protocols. As we know, training deep models effectively requires a large number of labeled datasets, but obtaining paired data of faces and skulls at a large scale is a challenging task. To address this challenge, we construct a paired dataset which consist of X-ray scans of the individual's faces and corresponding facial images from both front and lateral views. The purpose of conducting X-ray scans is that they are an affordable and widely accessible imaging technique that provides valuable information about skull structure. In addition, optical images of individuals' faces from lateral and frontal views are also taken. Finally, each scan is paired with a corresponding facial image of the volunteers who participated in the research. Figure \ref{fig:dataset} presents some samples of the pairwise dataset. To focus only on cranial structures (i.e., hard tissue of skull), the final dataset of 51 individuals is prepared by removing the soft tissue portions through a preprocessing step as mentioned in Section~\ref{soft_tissue_removal}.

 In the \emph{S2F} dataset, all volunteers are young adults, specifically aged between 21 and 30 years. Among the volunteers of 51 individuals, there were 22 females and 29 males. This diverse group provided valuable insights for our investigation into the relationship between X-ray imagery and facial morphology. The X-ray images were captured using a clinical X-ray scanner system at the institute's health centre. For training these generative models, we perform four types of data augmentation (i.e., horizontal flip, rotation, colour jitter and affine) to improve the quality of unbalanced \emph{S2F} datasets. Initially, we have 102 skull and face pairs, which include 51 lateral and 51 frontal views, respectively. The dataset is split into an 80:20 ratio for training and testing. Out of 102 pairs, 20 pairs are randomly selected for testing. For training the model, we performed the above-mentioned data augmentation on 82 training image pairs, resulting in a total of 410 pairs. Although augmentation increases the diversity of the training data, it does not fully replace the need for large-scale datasets. These generative frameworks are therefore evaluated under limited data conditions and further improvements in their performance can be achieved by incorporating larger and more diverse datasets. 

  \begin{figure}[!ht]
        \centering
        \includegraphics[width=1\linewidth, keepaspectratio,trim={7cm 4cm 6cm 5cm},clip]{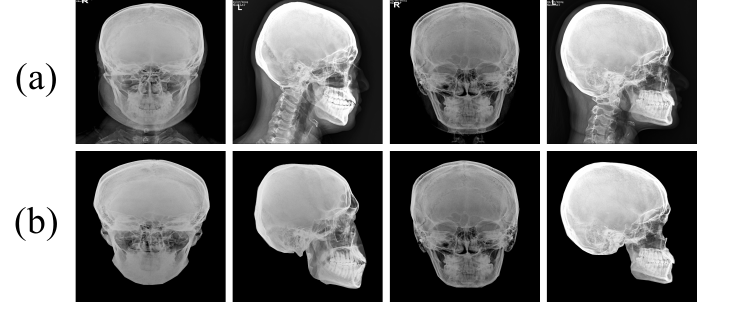}
        \caption{ Sample soft tissue eliminated X-ray images in frontal and lateral views, respectively, where (a) is the real X-ray images and (b) is the soft tissue eliminated images.}
        \label{fig:decomposed_pairs}
    \end{figure}
\section{Methodology}
\subsection{Ethics statement}
\begin{itemize}
    \item All methods in this study were carried out in accordance with relevant guidelines and regulations.
 \item All experimental protocols were approved by the Institute Ethical Committee (Human).
 \item Informed consent was obtained from all participants for both study participation and publication of identifying information in an online open-access publication.
\end{itemize}
 We investigate four generative models for the craniofacial reconstruction task. 
 The first one is CycleGAN~\cite{zhu2017unpaired}, the second is Conditioned GAN~\cite{isola2017image}, the third is CUT~\cite{park2020contrastive}, and the last one is FastCUT~\cite{park2020contrastive}.

 \begin{figure}[!ht]
        \centering
        \includegraphics[width=1\linewidth, keepaspectratio,trim={0cm 0cm 0cm 0cm},clip]{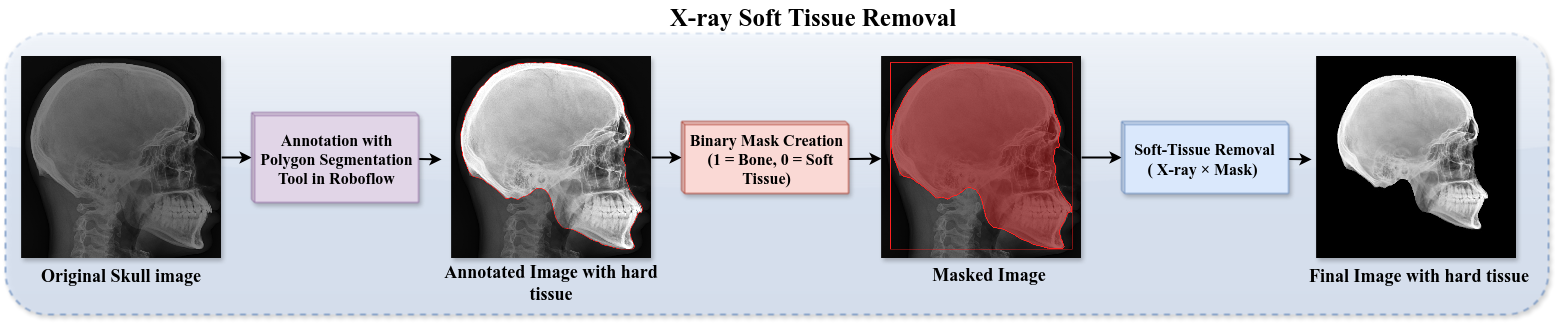}
        \caption{ Soft tissue removal process in which original x-ray image with soft tissue is first annotated with polygon segmentation tool and after that binary mask is created which is overlapped with original image to output hard tissue image.}
        \label{fig:soft_tissue_removal}
    \end{figure}
    \begin{table*}[t]
\centering
\caption{Nomenclature of symbols and notations used}
\label{tab:symbols}

\setlength{\tabcolsep}{5pt}

\begin{tabular}{llllll}
\toprule
Symbol & Description & Symbol & Description & Symbol & Description \\
\midrule

$X$ & Skull domain 
& $Y$ & Face domain
& $S$ & Skull features \\

$F$ & Face features
& $G_X,G_Y$ & Generators
& $D_X,D_Y$ & Discriminators \\

$\lambda_{cycle}$ & Cycle weight
& $\lambda_{id}$ & Identity weight
& $\tau$ & Temperature Parameter \\

FID & Fréchet Inception Distance
& IS & Inception Score
& SSIM & Structural Similarity Index \\

LPIPS & Perceptual similarity metric
& $H$ & Feature encoder
& $\mathbb{E}$ & Expectation Operator \\

\bottomrule
\end{tabular}
\end{table*}
\subsection{X-ray soft tissue elimination}\label{soft_tissue_removal} As mentioned in Section \S\ref{sec:dataset}, we have 2D X-ray images of faces that resemble skull images. However, the X-ray images also contain the effect of soft tissue. To ensure that the X-ray images accurately represent the skull, we manually removed the soft tissue components from the 2D X-ray images. Figure \ref{fig:decomposed_pairs} shows some samples of real and soft tissue eliminated X-ray images, respectively. Also, Figure~\ref{fig:soft_tissue_removal} shows the process of soft tissue removal. As raw X-ray images have soft tissue information, hence to obtain a skull which is only consists of hard tissue, we preprocess the original X-ray image. To standardize the preprocessing procedure across all samples, a consistent annotation protocol was followed for every X-ray image. First we annotated the X-ray image using polygon-based segmentation tool which is provided by Roboflow~\cite{roboflow}. During annotation, only hard tissue regions corresponding to the skull structure were selected, while soft tissue regions were excluded. To further ensure consistency, all annotations were reviewed under identical preprocessing conditions and image resolutions. Then, after annotating, we created a binary mask; 1 for bone and 0 for soft tissue. Finally, we multiply that mask with the original image, resulting in hard tissue skull image.

    \subsection{CycleGAN}
\begin{figure*}[!ht]
        \centering
        \includegraphics[width =17.5cm,keepaspectratio,trim={1cm 6.5cm 1cm 4cm},clip]{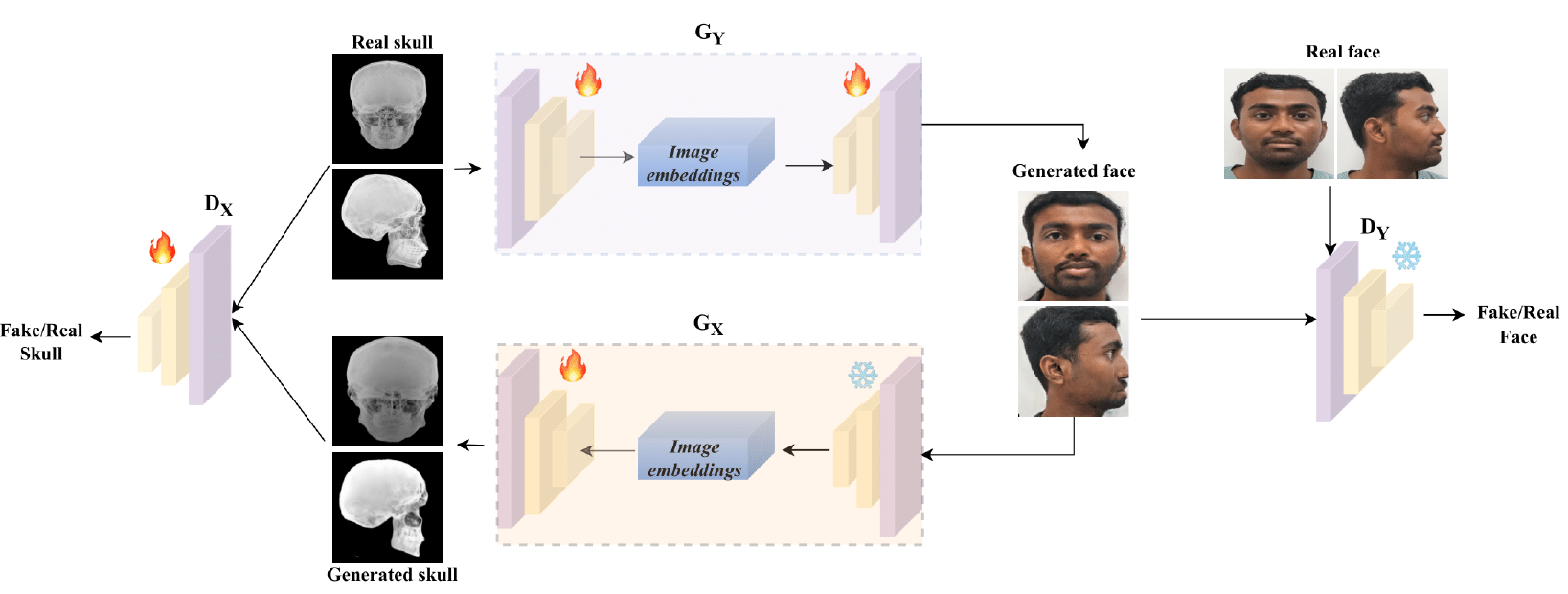}
        
        \caption{Our architecture consists of two frameworks, one is X-ray soft tissue decomposition, which decomposes soft tissue from the X-ray image so that it resembles the skull, and the other consists of two Generators. Generator G$_Y$ generates images for the first domain, which is \(skull\), while generator G$_X$ generates images for the second domain, which is \(face\). This design helps to produce images that look good and fit well with their context. Using both generators allows for better image creation in each domain. The discriminator D$_Y$ encourages the G$_Y$ to convert inputs from domain \(skull\) into outputs that are indistinguishable from those in domain \(face \). Conversely, discriminator D$_X$ serves the same purpose for \( face\), ensuring that generator G$_X$ translates inputs from domain \( face \) into outputs that closely resemble those in domain \( skull \). (Best viewed in colours)}
        \label{fig:framework}
    \end{figure*}

In this section, we present our first generative model, which is CycleGAN~\cite{zhu2017unpaired}, a generative framework as shown in Figure~\ref{fig:framework} for craniofacial reconstruction using 2D X-ray images. Initially, this model learns the mapping of embeddings across domains to perform reconstruction with unsupervised settings. Here, an unsupervised setting means that during training, CycleGAN does not need paired images of skull and face to learn the translation from the skull domain to the face domain. To enable learning without paired data, CycleGAN uses cyclic loss in Eq.~\ref{cyclic}, which ensure that when a skull image is translated to a face image, it must come back to the source domain. Also, to preserve the appearance of the face images that are already in target domain, this model uses identity loss in Eq.~\ref{identity}. All images were resized to 256 x 256.

\textbf{Adversarial loss}. In this work, we employ adversarial loss~\cite{goodfellow2014generative} to promote visual similarity between our output and images from the target domain, as follows:
\begin{equation}
\mathcal{L}_{\text{GAN}}(G, D, X, Y) = \mathbb{E}_{y \sim Y} [\log D(y)] + \mathbb{E}_{x \sim X} [\log(1 - D(G(x)))]
\end{equation}

\textbf{Cyclic loss in CycleGAN:}
\begin{equation}
\mathcal{L}_{\text{cycle}}(G_X, G_Y) = \mathbb{E}_{x \sim p_{\text{data}}(x)} \left[ \| G_X(G_Y(x)) - x \|_{1} \right] + \mathbb{E}_{y \sim p_{\text{data}}(y)} \left[ \| G_Y(G_X(y)) - y \|_{1} \right]
\label{cyclic}
\end{equation}

\textbf{Identity loss in CycleGAN:}
\begin{equation}
\mathcal{L}_{\text{identity}}(G_X, G_Y) = \mathbb{E}_{y \sim p_{\text{data}}(y)} \left[ \| G_Y(y) - y \|_{1} \right] + \mathbb{E}_{x \sim p_{\text{data}}(x)} \left[ \| G_X(x) - x \|_{1} \right]
\label{identity}
\end{equation}

Taking help from the above loss functions, we have defined cycleGAN loss as:
\begin{equation}
L(G_Y, G_X, D_Y, D_X) = L_{\mathrm{GAN}}(G_Y, D_Y, X, Y) + L_{\mathrm{GAN}}(G_X, D_X, Y, X) + \lambda_{cycle} L_{\mathrm{cycle}}(G_X, G_Y) + \lambda_{id}\mathcal{L}_{\text{identity}}(G_X, G_Y)
\label{eq:full_loss}
\end{equation}

where $\lambda_{cycle}$ and $\lambda_{id}$ are hyperparameters which controls the cycle consistency and identity preservation and are set to 10 and 5, respectively. Here, \emph{G$_X$}, \emph{G$_Y$} are generators for generating skull and face images respectively, and \emph{D$_X$}, \emph{D$_Y$} are discriminators for differentiating between the generated skull image and real skull and between generated face image and real face image respectively.
 In this craniofacial reconstruction, we represent \emph{X} as the skull and \emph{Y} as the face. As shown in Figure~\ref{fig:framework}, first, the input X-ray images are preprocessed to reduce the effect of soft tissue and enhance the underlying skull structures.

The framework of CycleGAN consists of two generators, \emph{G$_X$} and \emph{G$_Y$} which are responsible for mapping cross-domain representation between the skull (\emph{X}) and face (\emph{Y}) domains. Here, \emph{G$_Y$} learns the mapping function from skull-to-face (\emph{X} → \emph{Y}), while \emph{G$_X$} learns the inverse mapping (\emph{Y} → \emph{X}). The generator \emph{G$_X$} has a pretrained \emph{ResNet18} encoder, which is fine-tuned during training to extract meaningful structural features from generated face images. Whereas, the generator \emph{G$_Y$} employs a fully trainable encoder to learn facial feature representations from skull. This structure enables effective feature extraction in both domains while leveraging prior knowledge through transfer learning.
The use of separate generators for each domain helps in learning cross-domain feature more specifically, allowing the model to better capture the structural relationships between skull and facial representations.

To enhance the image generation process, two discriminators are embedded, \emph{D$_X$} and \emph{D$_Y$}. \emph{D$_Y$}'s role is to evaluate the output generated by \emph{G$_Y$}. Its goal is to ensure that the face images produced by \emph{G$_Y$} are so realistic and high-quality that they cannot be easily distinguished from actual face images. On the other hand, \emph{D$_X$} works with \emph{G$_X$} to ensure that the skulls it generates appear authentic and closely resemble real skulls.

By utilizing both generators and their corresponding discriminators, this architecture improves the quality of the images produced in each specific domain, leading to better overall image creation and a more accurate representation of both skulls and faces derived from X-ray images.

\subsection{Conditioned GAN}
In this second generative framework, we used Conditioned GAN~\cite{isola2017image}, also known as Pix2pix, to translate an image from the skull to the face domain. This model used a supervised setting, which means it needs paired images of the skull and their respective face to learn the mapping function. The generator in a conditioned GAN takes both a random noise vector (i.e., \emph{z}) and an input image (i.e., \emph{x}), and this input image guides the entire generator process. In Figure~\ref{fig:framework}, if we follow only one side of translation, then it can be modelled as the framework for a conditioned GAN, because this GAN has only one generator and one discriminator. Given below is the loss function used in the conditioned GAN:
\begin{equation}
\mathcal{L}_{cGAN}(G_Y, D_Y) = \mathbb{E}_{x,y} \big[ \log D_Y(x, y) \big] + \mathbb{E}_{x,z} \big[ \log (1 - D_Y(x, G_Y(x, z))) \big]
\end{equation}

\subsection{Learning cross-domain image representation using CUT}
In this section, we used the CUT~\cite{park2020contrastive} model, which is contrastive learning for unpaired image-to-image translation. This model is an improved version of CycleGAN in only one-sided image translation, basically consists of only one generator and one discriminator. Here, one-sided image translation means to translate an image from the skull to the face domain only, as our objective is to develop a mapping function that accurately relates the embeddings from skull features to their corresponding embeddings obtained from facial features. To accomplish this task effectively, several studies on a contrastive learning approach~\cite{bachman2019learning, chen2020simple,chen2020improved,park2020contrastive} have been done. This methodology is designed to minimize the distance between embeddings of similar pairs, where the skull and face belong to the same individual, while maximizing the distance between embeddings of dissimilar pairs, which belong to different individuals. By doing so, we enhance the model's ability to distinguish between unique facial characteristics tied to distinct skull structures, ultimately improving the precision of our mapping function.

\begin{figure}[!ht]
        \centering
        \includegraphics[width=1\linewidth, keepaspectratio,trim={8.5cm 8cm 1.5cm 3cm},clip]{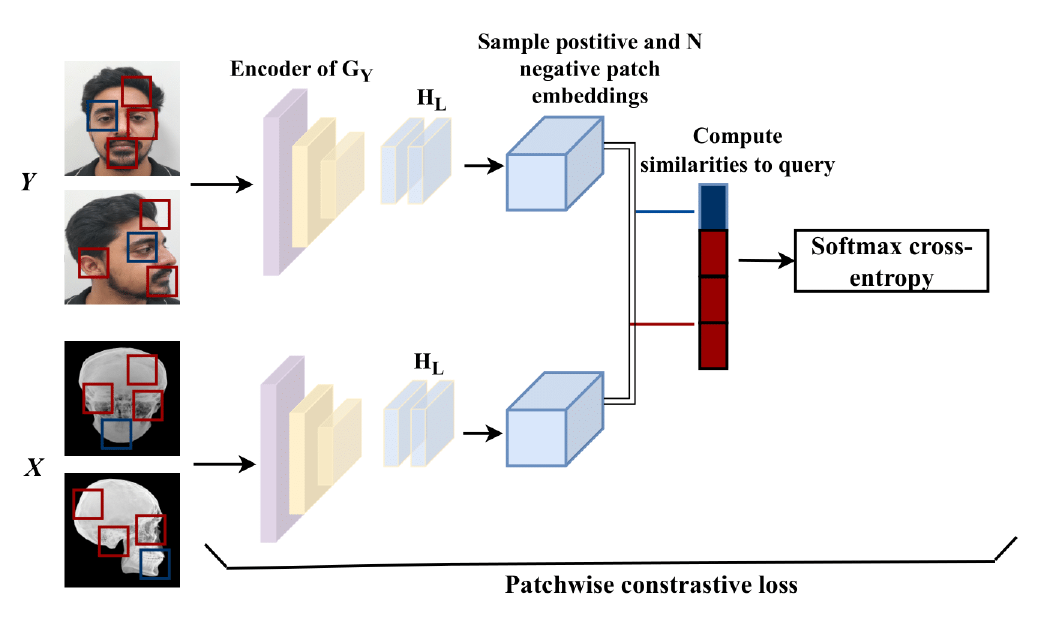}
        \caption{First, images from both domains (\emph{X} and \emph{Y}) are encoded using encoders of generator $G_Y$ and two-layer MLP networks ($H_L$). Then, the query patch from the output \emph{Y} is sampled and compared with the input patch \emph{X} at the same location, and finally, the cross-entropy loss function is used to find the patch-wise contrastive loss. (Best viewed in colors)}
        \label{fig:semantic}
    \end{figure}
    
\textbf{Patch-wise learning for craniofacial reconstruction.} Figure~\ref{fig:semantic} represents patch-wise learning for cross-domain image reconstruction. Instead of only working on the whole skull image, the CUT model ~\cite{park2020contrastive} also works on the patch level. Patch-wise loss used in the CUT model maps corresponding patches in the source and target domains at a specific location. For example, the nasal patch of the skull should more closely match the nasal patch of the face than the other patches of the given input skull. Hence, to make the model learn this type of patch-wise semantic attributes, we selected \emph{L} layers from the encoder part of CUT~\cite{park2020contrastive} model, extracted the feature maps from these layers, and then passed these feature maps through a small two-layer MLP network (\emph{H$_l$}). This MLP network produces a stack of features. Now, we have defined the stack of source (i.e., skull) features as:
\begin{equation}
    S = \{ H_{l}(\text{\emph{G}}^l_{enc}(x)) \}_{l=1}^{L}
\end{equation}
In the above equation, \emph{G$^{\text{l}}_{enc}$} is the output of the \emph{l-th} layer of the encoder of generator $G_Y$ where \(l \in \{1, 2, \ldots, L\}\) and \emph{x} is the skull image patch. 

In the same way, we encoded the target image (face) as:
\begin{equation}
     F = \{ H_{l}(\text{\emph{G}}^l_{enc}(G_Y(x))) \}_{l=1}^{L}
\end{equation}
 Now, we have encoded the skull (\emph{x}) and face (\emph{y}) image patches, and for patch-wise contrastive learning, we have used cross-entropy loss, which is given as:

\begin{equation}
\mathcal{\ell}(F, S^+, \{S^-\}) = -\log \left( \frac{\exp(F \cdot S^+ / \tau)}{\exp(F \cdot S^+ / \tau) + \sum_{n=1}^{N} \exp(F \cdot S^-_n / \tau)} \right)
\label{cross_entropy}
\end{equation}
Where,
\begin{itemize}
  \item $F$: generated face feature \(\in \mathbb{R}^K\) (\emph{K}-dimensional feature vectors from output \emph{face} patch)
  \item $S^+$: positive key  \(\in \mathbb{R}^K\) (from input \emph{skull} patch at the same location)
  \item $S^-$: negative keys  \(\in \mathbb{R}^{N \times K}\) (\emph{N} number of negative samples from other \emph{skull} patches)
  \item $\tau$: temperature hyperparameter which is set to 0.07.
\end{itemize}

The final patch-wise loss~\cite{park2020contrastive} for particular patch location ($s \in \{1, \ldots, S_l\}$) in a feature map is given as:

 \begin{equation}
\mathcal{L}_{\text{Patch}}(G_Y,H,X) = \mathbb{E}_{x \sim X} \Bigg[ \sum_{l=1}^{L} \sum_{s=1}^{S_l} \ell \big(F_{s}^{l}, S_{s}^{+l}, \{ S_{s}^{-l} \} \big) \Bigg]
\label{patch_loss}
 \end{equation}
Similarly, to preserve the identity of the target domain (face), generator of CUT model is provided with the input image from the target domain (i.e., face image). Hence, this process serves as a regularizer and prevent the generator from making unnecessary changes to the images that are already in the target domain (face) and for this identity loss in CUT is defined as:
 \begin{equation}
\mathcal{L}_{\text{identity}}(G_Y, Y) = \mathbb{E}_{y \sim Y} \, \| G_Y(y) - y \|_{1}
\label{patch_loss_1}
 \end{equation}

 And the final GAN loss with patch-wise loss function is given as:
\begin{equation}
\mathcal{L} = \mathcal{L}_{GAN}(G_Y, D_Y, X, Y) + \lambda_X \mathcal{L}_{\text{Patch}}(G_Y, H, X)+  \lambda_Y \mathcal{L}_{\text{identity}}(G_Y, Y) 
\label{final_loss}
\end{equation}

Where \emph{X} is the source domain image (i.e., skull), \emph{Y} is the target domain image (i.e., face), \emph{P} is the image patch and \emph{H} is the feature encoder, or the layers within G, that extracts feature maps for the Patch loss~\cite{park2020contrastive}. In practice, \emph{H} is implemented using selected intermediate layers of G, rather than as a separate network. $\lambda_X$ and $\lambda_Y$ are hyperparameters which are 1.
In the Eq.~\ref{final_loss}, if we put  $\lambda_X$ and $\lambda_Y$ equal to 10 and 0, respectively, then it can be modelled as FastCUT~\cite{park2020contrastive}. Table~\ref{tab:symbols} provides list of symbols and notations used in the paper.


\subsection{Face recognition (FR)  and image retrieval (IR) module for generated images.}\label{face_reognition}
\begin{figure}[!ht]
        \centering
        \includegraphics[width=1\linewidth, keepaspectratio,trim={1cm 0cm 1cm 0cm},clip]{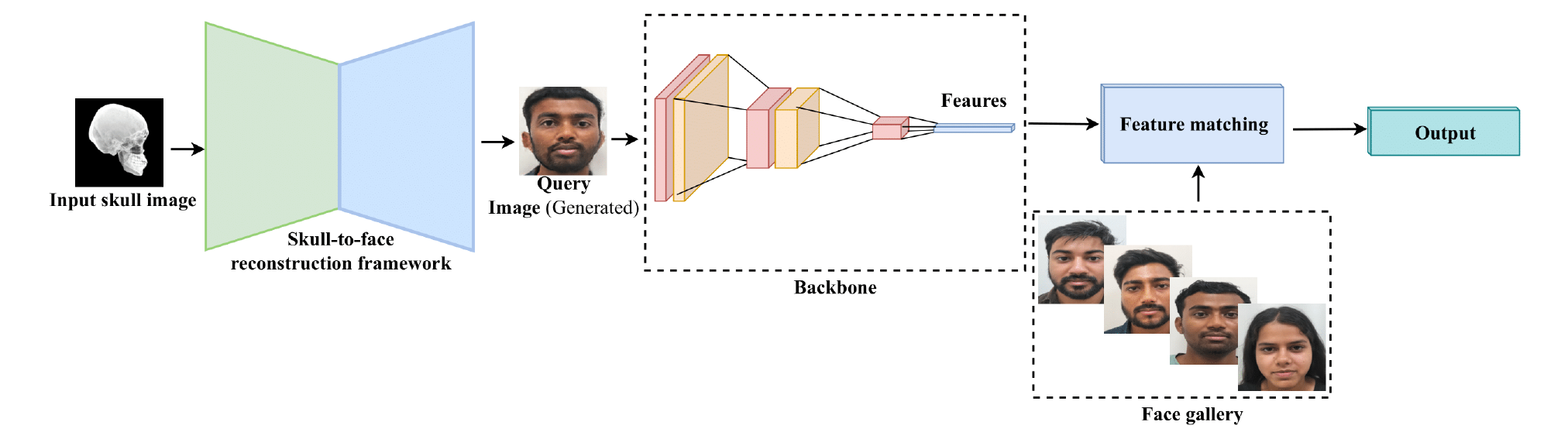}
        \caption{Face retrieval framework for the generated query face image from the gallery face images. First, face image is reconstructed from the input skull image with reconstruction framework, then this generated face is treated as query image and then feature of query face is extracted from the backbone which consists of different pre-trained deep models and then feature matching is done between the query feature and precomputed face gallery features. }
        \label{fig:detection}
    \end{figure}

In our experiments, to evaluate the effectiveness of the reconstructed faces generated by generative models, each generated face is compared to all real faces in a database, which consists of 200 face images. This face gallery database is collected from IITMandi student volunteers. For this, we utilized four deep learning models (VGG16, ResNet18, ResNet101 and DenseNet121) as a backbone to extract the features of generated images and compare them to the features of real faces in the database. A retrieval framework is essential for forensic applications because it enables efficient and accurate identification of unknown individuals through similarity-based searches in large databases. There were several studies conducted for image retrieval in many different applications, as mentioned in ~\cite{zaeemzadeh2021face, zhao2017memory, yu2020syntharch}. Figure~\ref{fig:detection} illustrates our proposed retrieval framework. When provided with a query image (generated face), the system retrieves the most relevant images (faces) from a gallery by matching the features of the query face image with the features of the gallery face images. In this framework, we have used four deep models~\cite{salimans2016improved,he2016deep,huang2017densely} to extract the features or embeddings of face images. Feature matching is performed by calculating the Euclidean distance between the embeddings of the query face image and the gallery face images. The face image from the gallery with the minimum distance to the query face is considered the matched face. We utilized two retrieval metrics—recall and mean average precision (mAP)—which are crucial in the field of forensics.

In forensic applications, recall and mAP are important retrieval metrics~\cite{manning2009introduction} because recall ensures that true identities are not missed, even if they do not appear as the top match, and mAP takes into account both the ranking of the results and the presence of relevant items, making it a comprehensive metric. Here, we calculated recall@k and mAP@k for top k retrieval. In forensic contexts, where multiple relevant images or identities may need to be retrieved, mAP effectively captures how well the system ranks all potential matches correctly while recall captures the availability of the identities in the top k retrieval.

\begin{figure}[t]
    \centering
    
    \begin{subfigure}[t]{0.48\linewidth}
        \centering
        \includegraphics[
            width=\linewidth,
            trim=12cm 2.5cm 12cm 2cm,
            clip
        ]{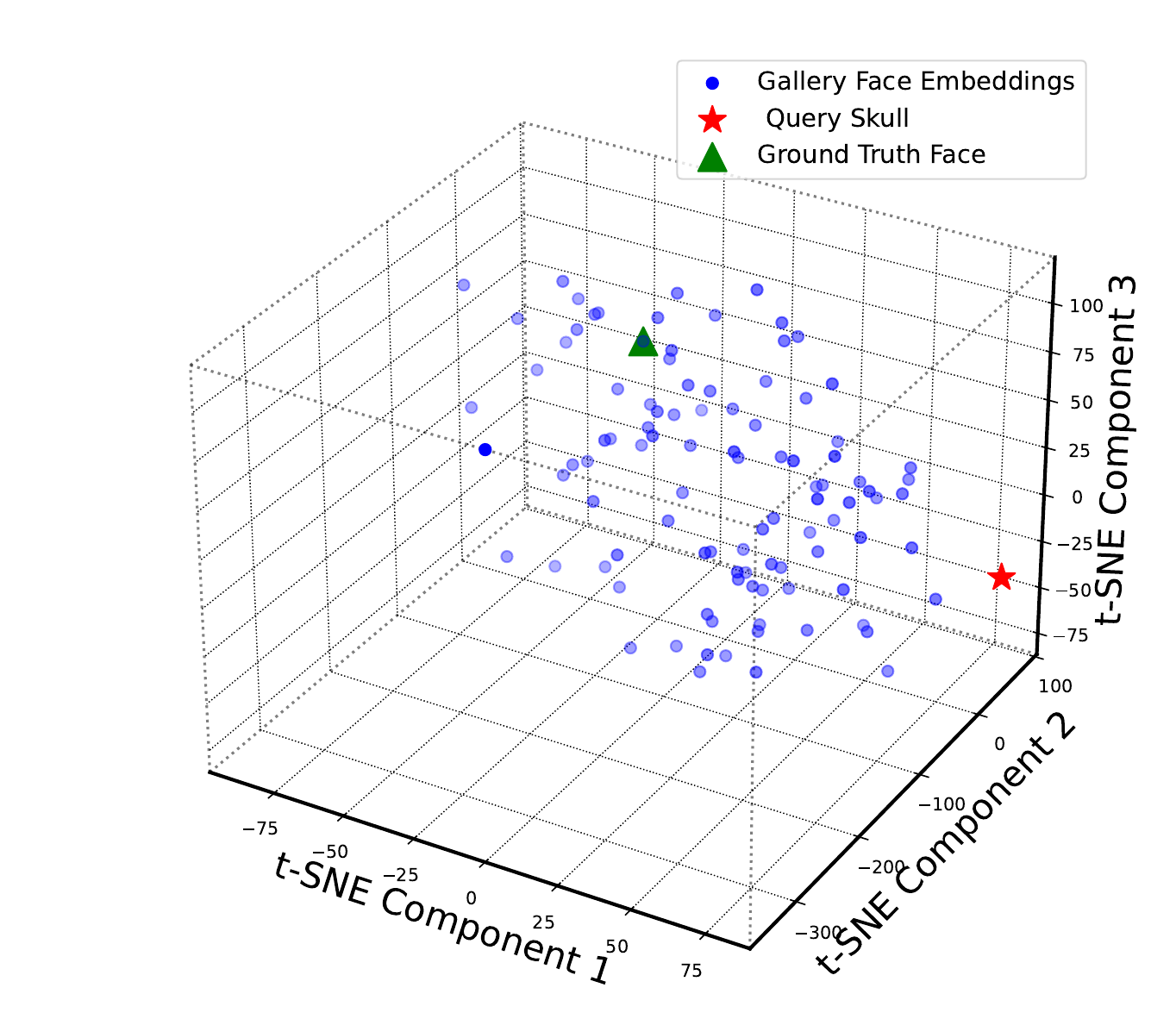}
        \caption{Before Training}
        \label{fig:tsne_before}
    \end{subfigure}
    \hfill
    \begin{subfigure}[t]{0.48\linewidth}
        \centering
        \includegraphics[
            width=\linewidth,
            trim=12cm 2.5cm 12cm 2cm,
            clip
        ]{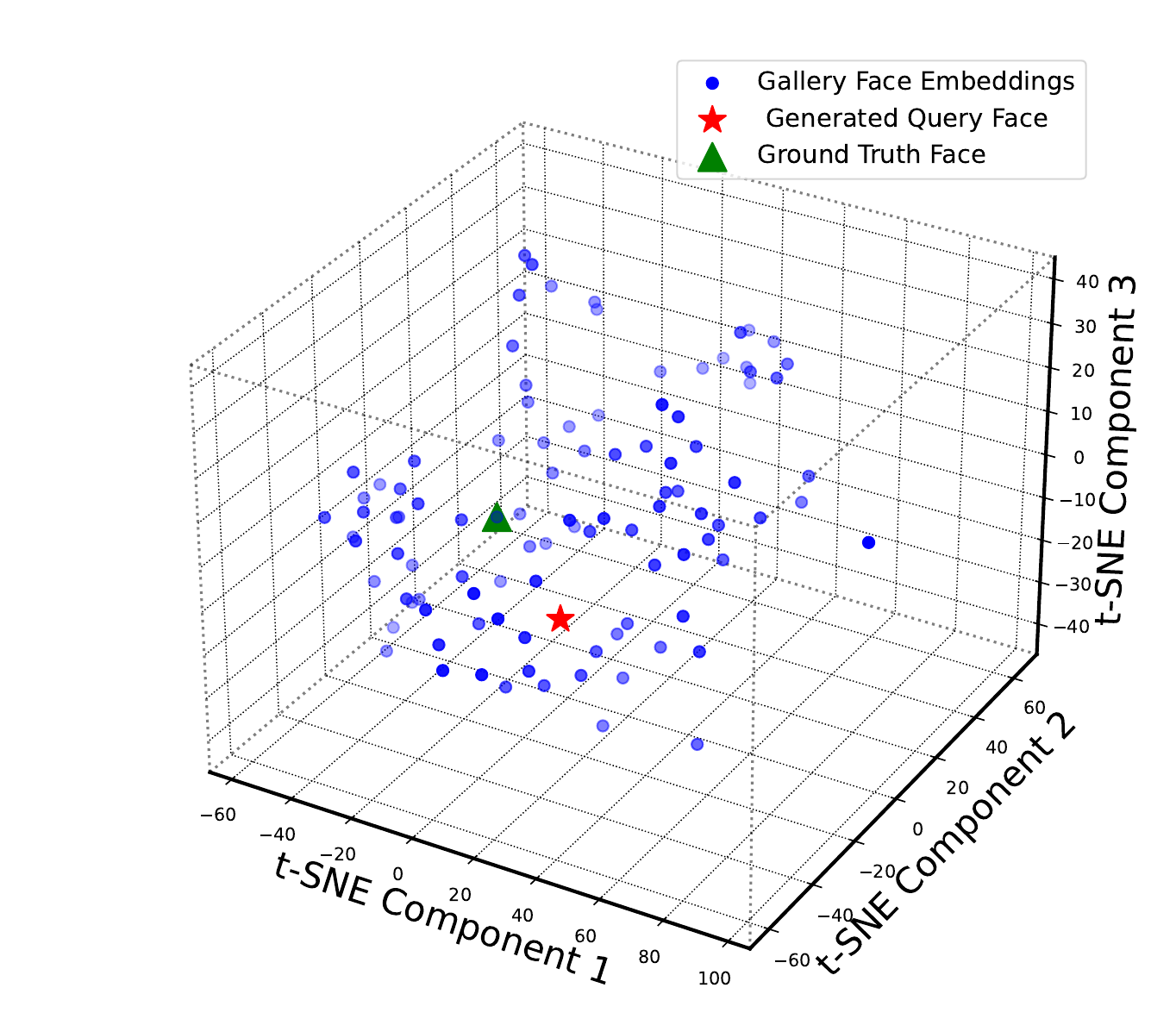}
        \caption{After Training}
        \label{fig:tsne_after}
    \end{subfigure}
    
    \caption{3D t-SNE visualization before and after training. Here, FastCUT generated images are used to plot with the ground truth and gallery face embeddings.}
\end{figure}

\section{Experiments}
The primary objective of the CR task is to identify the individual's face based on the underlying unknown skull; thus, the effectiveness of reconstructed faces in recognition algorithms is essential. Hence, we tested various baselines of generative models in unpaired and paired cross-domain image translation models with the benchmark \emph{S2F} dataset. 

 \textbf{Implementation details.} The proposed method is implemented using a 24GB Nvidia RTX A5000 GPU graphics card, designed for complex deep learning computations. We use transfer learning and finetune the encoder and decoder part of different generative models, for 500 epochs. We selected the best hyperparameter: a learning rate of 0.0001, a batch size of 16. For training these models with S2F dataset, we partitioned this dataset into training and test with the split ratio of 80:20 respectively. All images were resized to 256 x 256. We keep the same settings for training CycleGAN, cGAN and FastCUT as mentioned in~\cite{zhu2017unpaired,isola2017image, park2020contrastive}.
 
\subsection{Training with baseline models}
 There are a total of four generative models used. Table~\ref{tab:quantitative_full} shows the comparative analysis of various generative models using the \emph{S2F} dataset, and for each model, we have calculated five evaluation metrics to evaluate the quality of generated face image from the skull.

\subsection{Experimental results}

\textbf{Quantitative comparison.} To evaluate the performance of different generative models, we have used five evaluation metrics, including Fréchet Inception Distance (FID)~\cite{heusel2017gans}, Inception Score (IS )~\cite{salimans2016improved}, and Structural Similarity Index scores (SSIM)~\cite{wang2004image}, Learned Perceptual Image Patch Similarity (LPIPS)~\cite{zhang2018unreasonable} and ArcFace similarity~\cite{deng2019arcface}. These five metrics collectively assess generative quality, perceptual similarity, and identity preservation. From Table~\ref{tab:quantitative_full}, \begin{table}[ht]
    \centering
    \begin{subtable}[t]{0.95\linewidth}
\centering
\caption{Quantitative comparison of different methods for skull-to-face generation. Lower FID and LPIPS indicate better quality, while higher IS, SSIM, and ArcFace indicate better realism and identity preservation.}
\begin{tabular}{l|ccc|cc}
\toprule
\textbf{Methods}  
& \multicolumn{3}{c|}{\textbf{Image Quality}} 
& \multicolumn{2}{c}{\textbf{Identity \& Perceptual}} \\

\cmidrule(lr){2-4} \cmidrule(lr){5-6}

& FID$\downarrow$ & IS$\uparrow$ & SSIM$\uparrow$ 
& LPIPS$\downarrow$ & ArcFace$\uparrow$ \\

\midrule
CycleGAN~\cite{zhu2017unpaired} 
& 313.46 & 2.48 $\pm$ 0.14 & 0.33 & 0.5986 $\pm$ 0.0452 & 0.1091 $\pm$ 0.0716  \\

cGAN~\cite{isola2017image}      
& 148.84 & 1.66 $\pm$ 0.28 & 0.66 & \textbf{0.2660 $\pm$ 0.0659} & 0.2454 $\pm$ 0.1339 \\

CUT~\cite{park2020contrastive}  
& 305.04 & 2.67 $\pm$ 0.20 & 0.32 & 0.5605 $\pm$ 0.0718 & 0.1141 $\pm$ 0.0980 \\

\textbf{FastCUT}~\cite{park2020contrastive} 
& \textbf{63.65} & \textbf{2.72 $\pm$ 0.22} & \textbf{0.66} & 0.4281 $\pm$ 0.1563 & \textbf{0.5167 $\pm$ 0.0303} \\

\bottomrule
\end{tabular}
\label{tab:quantitative_full}
\end{subtable}

\newcommand{\cmark}{\ding{51}} 
\newcommand{\xmark}{\ding{55}} 

\begin{subtable}[t]{\linewidth}
\centering
\caption{Retrieval performance under different view settings and gallery sizes.}
\resizebox{\linewidth}{!}{
\begin{tabular}{l|cc|cccccccc|cccccccc}
\toprule
\textbf{Models} 
& \textbf{Front} & \textbf{Side}  
& \multicolumn{8}{c|}{\textbf{Gallery = 100}} 
& \multicolumn{8}{c}{\textbf{Gallery = 200}} \\

\cmidrule(lr){4-11} \cmidrule(lr){12-19}

& & 
& \multicolumn{4}{c|}{Recall@k} 
& \multicolumn{4}{c|}{mAP@k}
& \multicolumn{4}{c|}{Recall@k} 
& \multicolumn{4}{c}{mAP@k} \\

\cmidrule(lr){4-7} \cmidrule(lr){8-11}
\cmidrule(lr){12-15} \cmidrule(lr){16-19}

& & 
& 10 & 20 & 40 & 80 
& 10 & 20 & 40 & 80
& 10 & 20 & 40 & 80 
& 10 & 20 & 40 & 80 \\

\midrule
\multirow{3}{*}{VGG16}
& \cmark & \xmark  & 15.0 & 30.0 & 50.0 & 75.0 & 1.61 & 2.54 &3.27 & 3.95 & 5.00&15.0  &30.0 &45.0  &5.00  &5.51 &6.39  & 6.72\\
& \xmark & \cmark  & \textbf{23.5} & \textbf{44.1} & 55.8 & 88.2 & 6.67 & 8.11 &8.80 & 9.79 & 16.7 & 27.8 & 33.3 & 55.6 & 3.94 & 4.52 & 4.66 & 5.29 \\
& \cmark & \cmark  & 13.1 &23.6 &42.1  &76.3&4.75 &5.33 &6.03 &7.11 & 5.26 & 15.7 & 34.2 & 55.2 & 3.51 & 4.26 & 5.08 & 5.77 \\

\midrule
\multirow{3}{*}{ResNet18}
& \cmark & \xmark  & 10.0 & 30.0 & 45.0 & 90.0 & 3.12 & 4.74 &5.67 & 6.98 & 0.00 & 5.00 & 30.0 & 75.0 & 0.00 & 0.28 & 1.15 & 2.31 \\
& \xmark & \cmark  & 17.6 & 32.3 & \textbf{58.8} & 85.2 & \textbf{8.95} & \textbf{9.96} & \textbf{11.2} & \textbf{12.1} & 5.56 & 16.7 & 27.8 & \textbf{83.3} & 5.56 & 6.17 & 6.60 & 8.15 \\
& \cmark & \cmark  & 2.63 & 10.5 & 28.9 & \textbf{94.7} & 0.26 & 0.07 & 1.37 & 3.23 & 0.00 & 7.89 & 26.32 & 73.6 & 0.00 & 0.52 & 1.19 & 2.63 \\

\midrule
\multirow{3}{*}{ResNet101}
& \cmark & \xmark  & 15.0 & 20.0 & 40.0 & 75.0 & 7.17 & 7.62 & 8.37 & 9.37 & 5.00 & 15.0 & 25.0 & 55.0 & 0.83 & 1.62 & 1.99 & 2.67 \\
& \xmark & \cmark  & 27.7 & 27.8 & 44.4 & 94.4 & 8.06 & 8.06 &  8.65 & 9.95 & \textbf{27.7} & \textbf{27.8} & 38.9 & 72.2 & \textbf{7.22} &\textbf{ 7.22} & \textbf{7.52} & \textbf{8.44} \\
& \cmark & \cmark  & 10.5 & 18.4 & 31.5 & 68.4 & 4.02 & 4.58 &5.16 & 6.15 & 10.5 & 15.7 & 26.3 & 55.2 & 4.02 & 4.44 & 4.88 & 5.60 \\

\midrule
\multirow{3}{*}{\textbf{DenseNet121}}
& \cmark & \xmark  & 5.00 & 15.0 & 35.0 & 80.0 & 1.00 & 1.85 & 2.75 & 3.73 & 5.00 & 10.0 & 25.0 & 55.0 & 1.00 & 1.53 & 2.08 & 2.92 \\
& \xmark & \cmark  & 11.1 & 22.2 & 44.4 & 88.9 & 2.78 & 3.68 &4.49 &5.58 & 11.1 & 22.2 & 38.8& 72.2& 2.78 & 3.59 & 4.16 & 5.12 \\
& \cmark & \cmark  & 18.4 & 23.6 & 47.3 & 86.8 & 5.33 & 5.78&6.85 &7.88 & 15.7 & 21.0 & \textbf{39.4} & 73.6 & 4.6 & 5.18 & 5.94 & 6.95  \\

\bottomrule
\end{tabular}
}
\label{tab:view_gallery_topk}
\end{subtable}
    \caption{(a) Quantitative analysis of various methods using the \emph{S2F} dataset with FID, IS, and SSIM metrics.  
             (b) Comparison of recall and mAP at different value of k (i.e., k = 10, 20, 40 and 80) and also with two gallery size (i.e., 100 and 200) for different face recognition models. Here, FastCUT generated face images are used for retrieval.
             Best results are shown in bold.}
    \label{tab:combined_retrieval}
\end{table}
FastCUT model has shown the best performance in terms of FID and IS, which indicates better generative quality and distribution alignment than other models. Additionally, FastCUT attains the highest ArcFace similarity, suggesting improved identity preservation when compared to other generative models. However, cGAN achieves the lowest LPIPS score, indicating better perceptual similarity to ground-truth images. It is also important to note that FID and IS are standard metrics for evaluating quality of generated image in generative models. However, these metrics are designed to compute over large image samples. But, due to the limited test set size (20 samples), these values may not be statistically stable and are reported primarily for relative comparison. Therefore, additional metrics such as SSIM, LPIPS, and ArcFace similarity are used to provide a more reliable evaluation.  Table~\ref{tab:view_gallery_topk} provides a comparative analysis of face retrieval performance across different models for varying values of k and gallery sizes. Here, results on retrieval at different values of k and gallery size show high recall@k but low map@k. This indicates that when correct identity is retrieved, it is not consistent ranked at the top positions. Hence, from this we can conclude that the generated face images preserve sufficient identity but lack the discriminative attributes required for accurate matching and ranking.

\begin{figure}[ht]
    \centering
    \begin{subfigure}{0.48\linewidth}
        \centering
        \includegraphics[width=1\linewidth, keepaspectratio,trim={5.5cm 1.5cm 4.3cm 1.3cm},clip]{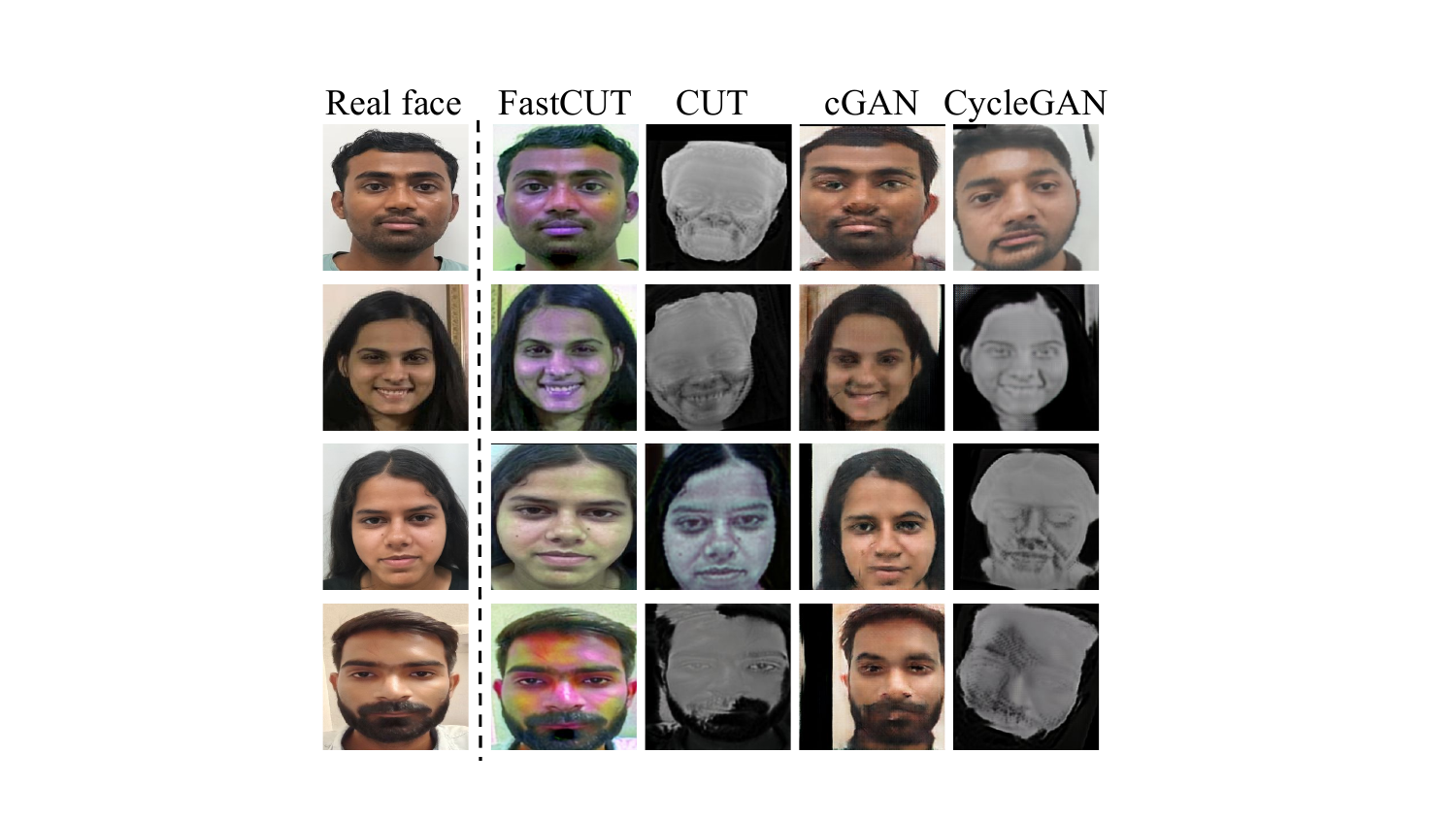}
        \caption{Qualitative comparison, where the generated \emph{front face} is compared with the ground truth \emph{front face} when translating from skull to face.}
        \label{fig:qualitative_1}
    \end{subfigure}
    \hspace{0.02\linewidth}
    \begin{subfigure}{0.48\linewidth}
        \centering
        \includegraphics[width=1\linewidth, keepaspectratio,trim={5.5cm 1.5cm 4.3cm 1.3cm},clip]{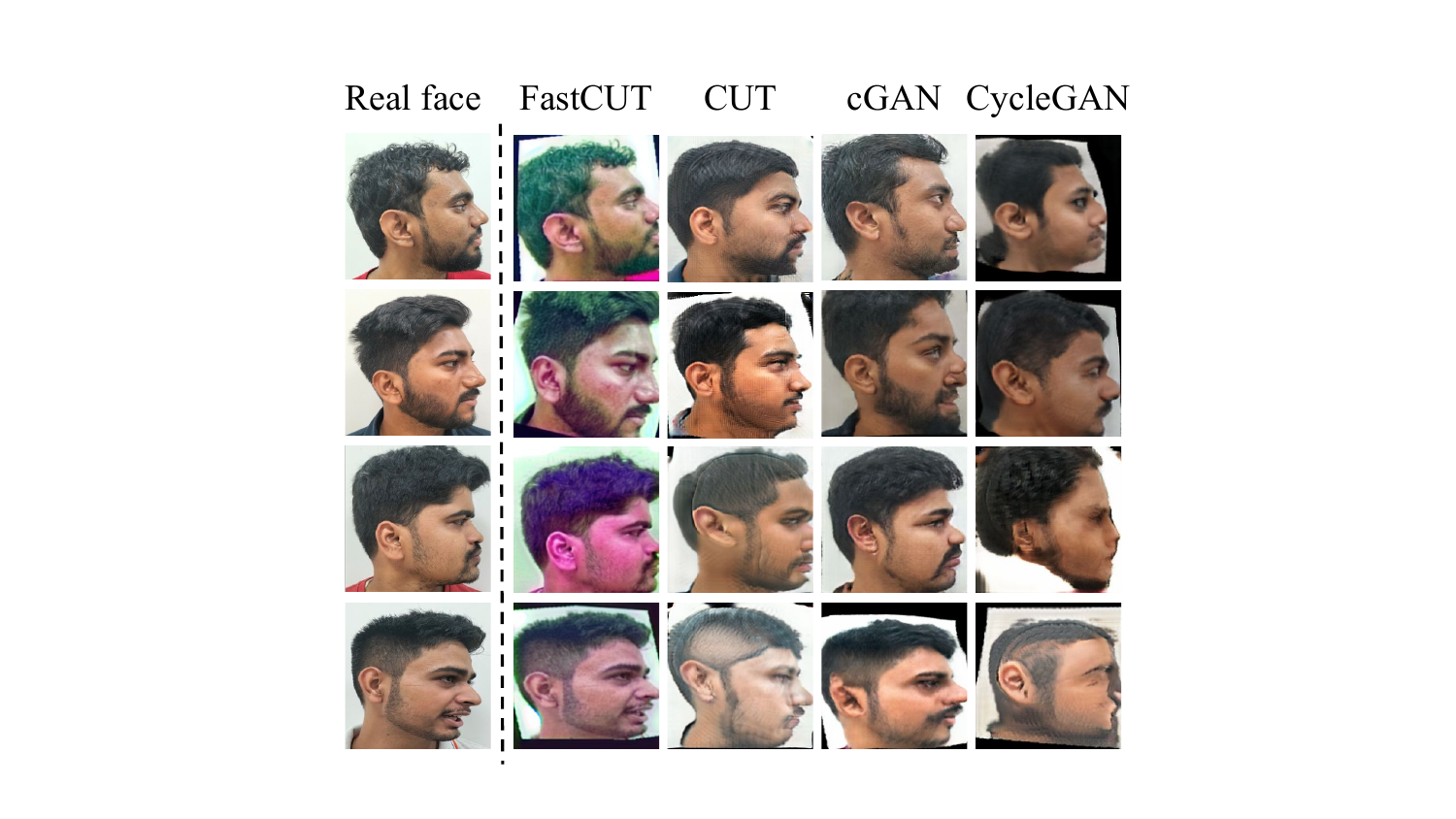}
        \caption{Qualitative comparison, where the generated \emph{side face} is compared with the ground truth \emph{side face} when translating from skull to face.}
        \label{fig:qualitative_2}
    \end{subfigure}
    \caption{Overall qualitative comparison between different generative models in (a) and (b).}
    \label{fig:qualitative}
\end{figure}

\begin{figure}[!ht]
        \centering
        \includegraphics[width=1\linewidth, keepaspectratio,trim={3cm 2.5cm 2cm 1cm},clip]{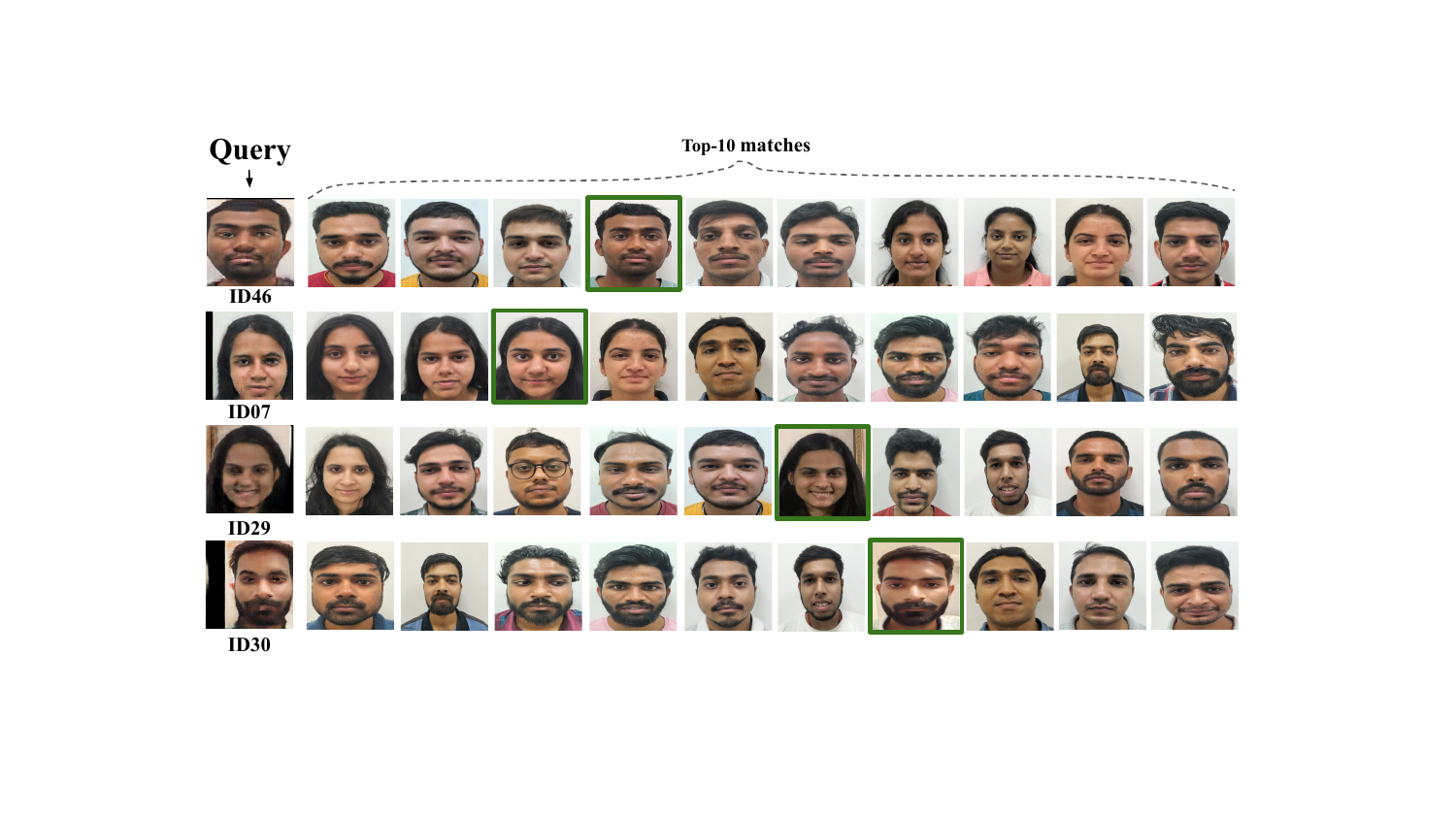}
        \caption{Top-10 retrieval results in which the generated image is treated as the query image and gallery face image with a green border is the correct face.}
        \label{fig:retrieval_query}
    \end{figure}

\textbf{Qualitative comparison.}
Figure~\ref{fig:tsne_before} shows 3D visualization of embeddings of query skull, ground truth face with the gallery face embeddings before generation, whereas Figure~\ref{fig:tsne_after} shows 3D distribution of embeddings of ground truth face with the gallery faces images and generated face using the FastCUT model. Here, we have used pretrained ResNet18 backbone to extract the embeddings, which demonstrates improved clustering of generated faces toward their corresponding ground-truth identities. Figure~\ref{fig:qualitative_1} and Figure~\ref{fig:qualitative_2} present qualitative comparisons of skull-to-face translation results for both lateral and frontal views across different generative models. Among the evaluated methods, FastCUT produces more structurally coherent and semantically consistent facial representations with \emph{S2F}dataset. Additionally, Figure~\ref{fig:retrieval_query}  illustrates retrieval results for the top-10 gallery matches, where correct identities are shown in a green border. Here also we have utilized pretrained ResNet18 model for retrieval. The results demonstrate that improved identity preservation in generated faces contributes positively to retrieval performance.

    
    
    
    
\begin{figure*}[!ht]
\centering

\begin{minipage}[c]{0.58\linewidth}
\centering

\includegraphics[
width=\linewidth,
keepaspectratio,
trim={4cm 3cm 9.5cm 1cm},
clip]{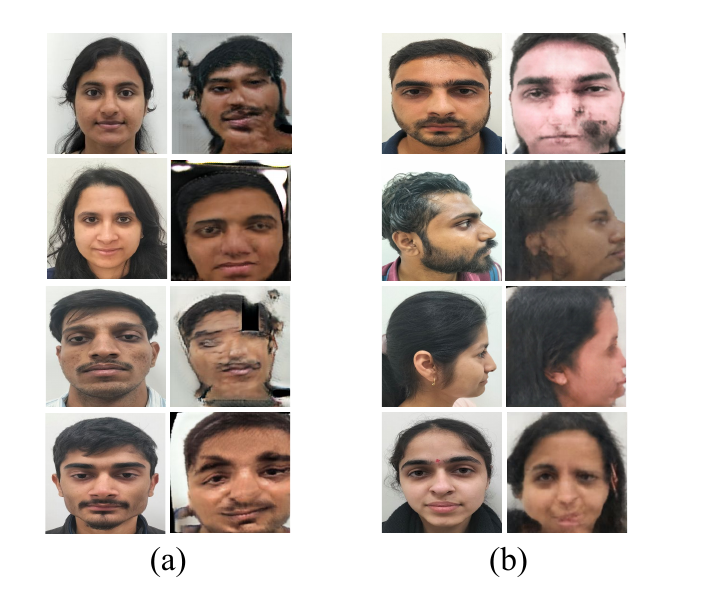}

\captionof{figure}{Failed case of two different generative models. Here, (a) shows the failed case of CycleGAN, whereas (b) shows the failed case of FastCUT when translating skull$\rightarrow$face images.}

\label{fig:failed_case}

\end{minipage}
\hfill
\begin{minipage}[c]{0.38\linewidth}
\centering

\captionof{table}{Qualitative comparison of generative models based on visual quality (VQ), identity preservation (IP), and stability. FID, IS, and LPIPS correspond to VQ, ArcFace corresponds to IP, and standard deviation represents stability. (Mod refers to Moderate).}

\label{tab:qualitative_comparison}

\vspace{0.2cm}

\begin{tabular}{lccc}
\toprule
Model & VQ & IP & Stability \\
\midrule
CycleGAN & Low & Low & Low \\
cGAN     & Mod--High & Mod & High \\
CUT      & Mod & Low & Mod \\
\textbf{FastCUT} & \textbf{High} & \textbf{High} & \textbf{High} \\
\bottomrule
\end{tabular}

\end{minipage}

\end{figure*}

\section{Discussion \&\ Limitations}
From this work, we conducted proof-of-concept towards craniofacial identification using different generative models. This study faces challenge, due to the absence of publicly available benchmark paired skull-face image datasets. As a result, we have been unable to assess the performance of different generative models against public craniofacial datasets. However, as more diverse and large-scale datasets become available, these generative models can be further fine-tuned and rigorously evaluated, leading to more robust and generalizable results. Our dataset, IITMandi\_S2F consists of Indian individuals with age group (21–30 years). It was collected under controlled conditions to ensure consistency and accuracy. This controlled environment allows us to deliver high-quality data for initial training and validation, while such controlled acquisition improves the experimental setup, but it also limits the model's generalizability. Consequently, these generative models may learn population specific characteristics. But in real world forensic scenarios, we may not always provide such ideal conditions due to noisy image acquisition protocols, which can compromise the quality of the images used for identification purposes. To partially address these issues, we have implemented different data augmentation techniques as mentioned in section 3. These techniques enhance the robustness of these generative models by adding variations in pose, lighting, and scale, which are common in real-world applications.

Despite these efforts, we recognize that further enhancements are necessary to develop a framework that can withstand the real world conditions which are encountered in actual forensic investigations. Hence, our future work will focus on two key areas: expanding the size of our dataset by incorporating a more diverse range of samples in terms of age, sex and data acquisition with different machines and improving the overall robustness of these generative models. We also observed that generative models like GANs are very good in the image generation task, but there are several failure cases when these models fail to generate a realistic image similar to the ground truth image. Figure~\ref{fig:failed_case} shows several failure cases. The reason for this failure case might be due to the complex distribution of the training dataset and very high cross-modality gap between the skull and face images. We also explored that with the \emph{S2F} dataset, the heavy architecture of some GAN models, like CycleGAN, fails to capture the intrinsic properties of skull images with very low SSIM (0.33) and ArcFace score (0.1091 $\pm$ 0.0716), while light weight model like FastCUT effectively captures the identity-related deep features from skull with high SSIM (0.66) and ArcFace score (0.5167 $\pm$ 0.0303). Table~\ref{tab:qualitative_comparison} provides the qualitative comparison of different generative models based on three parameters. Visual Quality (VQ), Identity Preservation (IP) and Stability. VQ is primarily inferred from FID, IS, and LPIPS scores, where lower FID/LPIPS and higher IS indicate more realistic image generation. IP is evaluated using ArcFace similarity, which measures the preservation of discriminative facial characteristics. Stability reflects the consistency of generated outputs across different samples and is assessed through variance in quantitative metrics.
\subsection{Conclusion}
This work presents a comparative study on the different generative methods for craniofacial reconstruction using 2D frontal and lateral-view images from facial scans. We conducted experiments on 2D X-ray images because they are easier to collect than other imaging modalities (i.e., CT, MRI). The results demonstrate that these generative techniques can be valuable in forensics, particularly when other identification methods (such as fingerprinting, dental records, radiological materials, or DNA analysis) are ineffective for human remains. Through comparative quantitative and qualitative analyses of various generative models for the skull-to-face image translation problem, it is observed that the FastCUT model outperforms other generative models in cross-domain image translation. This model can generate realistic human faces from the provided 2D X-ray skull images as reflected by FID, IS and ArcFace scores. However, cGAN model achieve better perceptual similarity ( LPIPS), which indicated the trade-off between visual representation and identity preservation. Despite of these promising results, these generative models constrained by dataset size and demographic diversity, which limits their ability to learn robust and generalizable skull-to-face mappings. We have also conducted experiments on the face retrieval task by matching the features of the query
face image (i.e., generated image) with the features of gallery face images, and the results show that the correct identity is often
retrieved within the tok-k (high recall@k), although the ranking performance (mAP@k) remains limited. We state that the current framework is more suitable for candidate retrieval rather than precise identity ranking, and improving identity preservation remains an important direction for future work. Overall, this study demonstrates the potential of these generative models for forensic craniofacial identification.

\section{\textbf{Data availability}}
 Our data will be made available on request.    Additionally, the S2F dataset used in this study, which includes X-ray images and their corresponding face vectors (vector representation of face images), is publicly available at: \url{https://github.com/singh-ml/IIT_Mandi_S2F}.
 However, the original face image data cannot be shared due to privacy concerns.
\section{\textbf{Funding}} This research is funded by seed grant project No. IITM/SG/DIS-ROS-SPA/111, Indian Institute of Technology Mandi, Department of Higher Education, Ministry of Education, Government of India.

\section{\textbf{Author Contribution declaration }}
Conceptualization, Ravi Shankar Prasad; Methodology, Ravi Shankar Prasad and Dinesh Singh; Data curation, Ravi Shankar Prasad; Original Draft Preparation, Ravi Shankar Prasad; Review \& Editing, Ravi Shankar Prasad and Dinesh Singh.
\section{Acknowledgment}
 We want to sincerely thank the volunteers who dedicated their time and effort to help create the benchmark dataset called \emph{S2F}. We also appreciate the cooperation and assistance from the Institute Health Centre during the data collection process and IIT Mandi for necessary support to this research. During the preparation of this work the author(s) used Grammarly in order to correct the spelling and grammar. After using this tool, the author(s) reviewed and edited the content as needed and take(s) full responsibility for the content of the publication.

\bibliography{sample}

\end{document}